\documentclass[letterpaper, 10 pt, journal, twoside]{IEEEtran}
\usepackage[utf8]{inputenc}
\usepackage{array}
\usepackage[hidelinks]{hyperref}
\usepackage[pdftex]{graphicx}
\usepackage[mode=buildnew]{standalone}
\usepackage{tikz}
\usepackage{booktabs}
\usepackage{siunitx}
\usepackage{amsmath}
\usepackage{amsfonts}
\usepackage{flushend}
\usepackage{bbold}
\usepackage{caption}
\usepackage{cite}
\usepackage{subcaption}
\usepackage{url}

\newcommand{\mdiff}[1]{#1}

\begin{document}
\title{
An Outlier Exposure Approach to Improve Visual Anomaly Detection Performance for Mobile Robots.
} 

\author{Dario Mantegazza, Alessandro Giusti, Luca Maria Gambardella and J\'er\^ome Guzzi
\thanks{This work was supported by the Swiss National Science Foundation (SNSF) through the NCCR Robotics and by the European Commission through the Horizon 2020 project 1-SWARM, grant ID 871743.} 
\thanks{Dario Mantegazza, Alessandro Giusti, Luca Maria Gambardella and J\'er\^ome Guzzi are with the Dalle Molle Institute for Artificial Intelligence (IDSIA), USI-SUPSI, Lugano, Switzerland. (email:
{\tt\footnotesize dario.mantegazza@idsia.ch; alessandrog@idsia.ch; luca@idsia.ch; jerome@idsia.ch})
}
\thanks{Supplementary materials (code and
dataset) are available at \url{https://github.com/idsia-robotics/hazard-detection}.}
}
\date{October 2022}

\maketitle

\thispagestyle{empty}
\pagestyle{empty}

\begin{abstract}
We consider the problem of building visual anomaly detection systems for mobile robots.  Standard anomaly detection models are trained using large datasets composed only of non-anomalous data.  However, in robotics applications, it is often the case that (potentially very few) examples of anomalies are available.  We tackle the problem of exploiting these data to improve the performance of a Real-NVP anomaly detection model, by minimizing, jointly with the Real-NVP loss, an auxiliary outlier exposure margin loss.  We perform quantitative experiments on a novel dataset (which we publish as supplementary material) designed for anomaly detection in an indoor patrolling scenario. On a disjoint test set, our approach outperforms alternatives and shows that exposing even a small number of anomalous frames yields significant performance improvements.
\end{abstract}


\section{Introduction}
\label{sec:intro}

\IEEEPARstart{I}{n} many applications, mobile robots are expected to operate autonomously in an unknown, uncontrolled environment perceived through onboard cameras. While designing the robot’s perception system, one explicitly accounts for potential hazards that the robot could face: for example, ground robots can typically perceive and avoid obstacles or uneven ground. However, to achieve long-term unattended operation, it is important to endow the robot with the ability to detect \emph{unexpected} situations, which are potential hazards and possibly very rare, that were not explicitly considered during system design.  For example, a ground robot performing indoor patrolling tasks might be faced with smoke, loss of ambient lighting, a wet floor, a dirty camera lens, a misalignment in the camera’s geometry, and other situations that can not be enumerated in advance. Because one does not have any model of how all these different types of anomalies may appear, an attractive approach is to consider anything that is novel or unusual (as compared to the robot’s normal experience) as a potential hazard to be detected and possibly reported to human supervisors.  This problem is known as Anomaly Detection or Out-of-distribution Detection in the Machine Learning literature, and several recent works \cite{wellhausen2020safe, khalastchi2015online, mantegazza2021sensing, blum2021fishyscapes} propose its integration into robot perception pipelines.

Anomaly Detection techniques assume that a large amount of non-anomalous data is available (e.g., data from past patrolling missions); only these data are used to train a model.  For a given new observation (e.g., a frame acquired by the robot camera), the model outputs an anomaly score, that quantifies how likely it is that the observation comes from a different distribution than the learned distribution of normal data.  In real-world robotics scenarios, this approach is difficult to apply because normal data has a large variability
and relevant anomalies might affect the image data in subtle ways (e.g., a puddle of water on the floor only affects a small part of the image). Absent a prior on the expected appearance of relevant anomalies, it is difficult to learn anomaly detection models that perform well enough to be useful in practice.  

However, in most realistic use-cases one can expect that at least some examples of anomalies will be available when training the anomaly detection model, thus extending beyond the pure anomaly detection paradigm. For example, \mdiff{samples} of possible anomalies might be collected during system design, and others after robot deployment and with partial human supervision. These examples might not represent all possible types of anomalies, and might be available in much smaller quantities than normal data.  Still, they represent useful pieces of information that should be integrated into the model.  \mdiff{To handle these additional training data, one baseline solution is to use the resulting dataset, composed of many normal and few anomalous samples, to train a binary classifier}.  This approach is suboptimal because the few available examples of anomalies do not cover the space of all possible anomalies. 

The \textbf{main contribution} of this paper is methodological: we combine Real-NVP~\cite{dinh2016density}, a state-of-the-art anomaly detection approach, with the concept of \emph{Outlier Exposure}, which was recently proposed in the deep learning literature~\cite{hendrycks2018deep} to train more robust anomaly detectors by providing large amounts of examples of outliers, sampled from completely different datasets than the dataset of normal samples.  In our context, we instead \mdiff{assume that} the exposed anomalies are few and correspond to relevant anomalies for the robot. This is implemented by using an additional margin loss term during training of a Real-NVP model; this additional term enforces that the exposed anomalies are assigned higher anomaly scores than random normal samples, but does not imply any expectation that the exposed anomalies are representative of all anomalies. We evaluate the method on a new real-world dataset, which we release as supplementary material as a \textbf{secondary contribution}, acquired by a ground patrolling robot;
experiments show that exposing even just a few anomalous frames significantly improves performance, comparing favorably with alternative approaches. 



\section{Related Work}
\label{sec:related}
\mdiff{The anomaly detection} literature spans a large variety of applications, including medical imaging~\cite{schlegl2017unsupervised}, industrial manufacturing inspection~\cite{scime2018multi, haselmann2018anomaly}, surveillance~\cite{chakravarty2007anomaly}, autonomous robot navigation~\cite{wellhausen2020safe}, fault detection~\cite{khalastchi2015online}, intrusion detection~\cite{birnbaum2015unmanned}, and agriculture~\cite{christiansen2016deepanomaly}.
Ruff et al.~\cite{ruff2021unifying} define anomalies as observations that deviate considerably from a concept of normality; Chandola et al.~\cite{chandola2009anomaly} state that anomalies are \mdiff{``}patterns that don't conform to expected behavior\mdiff{''}. 
In the field of robotics, anomalies can be detected on readings from exteroceptive~\cite{wellhausen2020safe, christiansen2016deepanomaly}, or proprioceptive~\cite{khalastchi2015online, birnbaum2015unmanned} sensors.

We focus our analysis on exteroceptive visual sensors like cameras.
On image data, anomalies can be classified~\cite{ruff2021unifying} as either \emph{low-level} -- i.e., related to image appearance, such as brightness, noise, blur -- or \emph{high-level} -- i.e., involving anomalies in the data semantics, such as an open pothole in an environment where potholes are always closed.  Detecting the latter is harder, and requires approaches capable of extracting and representing semantic information from images.

\subsection{Models and Approaches}
\paragraph*{Reconstruction-based methods}
A standard deep learning approach to visual anomaly detection uses undercomplete autoencoders~\cite{kramer1992autoassociative,cho2014learning}.
The typical architecture of autoencoders consists of two modules, the encoder, and decoder, separated by a bottleneck. The encoder and decoder are convolutional networks that, respectively, encode the input image into a low-dimensional embedding and decode the high-level informative embedding into an image.  The bottleneck limits the embedding size thus inducing a reproduction error in the output. This error can be exploited to detect anomalies when an autoencoder is trained only on normal images. Such a model, when tasked to reproduce an image that contains an anomaly, will fail to correctly reconstruct it; the reconstruction error can therefore be used as an anomaly score for the given input.
\paragraph*{Normalizing Flows}
Normalizing Flow models are used both for generative purposes~\cite{abdelhamed2019noiseflow,kingma2018glow,ho2019flow++} and in the context of anomaly detection~\cite{wellhausen2020safe,blum2021fishyscapes}.
These models \mdiff{rely on coupling layers, 
invertible transformations} that map the inputs to a chosen latent distribution, such as a normal distribution. 
For anomaly detection, one \mdiff{uses} the learned mapping to directly estimate the likelihood of a sample with respect to the distribution of non-anomalous samples used for training.
In this paper, we use Real-NVP~\cite{dinh2016density}, a type of Normalizing Flow model.

\paragraph*{Outlier Exposure}
Our approach combines Real-NVP with the concept of Outlier Exposure~\cite{hendrycks2018deep}: training anomaly detectors against an auxiliary dataset of known outliers.  In the existing literature, these known outliers are sampled from large benchmark datasets (e.g., 80 Million Tiny Images~\cite{torralba200880} or ImageNet-22K~\cite{deng2009imagenet}), which contain a huge amount of data that is semantically very different than in-distribution data; in this case, the approach helps the model to generalize and detect unseen anomalies.  Anomaly detection tasks on real-robot datasets differ, since much smaller datasets are available, and anomalies can be semantically very similar to normal images.


\subsection{Anomaly Detection In Robotics}

\paragraph*{On low-dimensional data}
A large amount of literature deals with detecting anomalies using low-dimensional (but potentially high-frequency) data streams from proprioceptive or exteroceptive sensors.  On this type of data, Khalastchi et al. rely on simple metrics such as the Mahalanobis Distance with good results in multiple scenarios~\cite{khalastchi2011online, khalastchi2015online}, whereas Sakurada et al.~\cite{sakurada2014anomaly} compare autoencoders to PCA and kPCA for spacecraft telemetry data.
Birnbaum et al.~\cite{birnbaum2015unmanned} propose an approach for detecting cyber attacks, sensors faults, or structural failures on Unmanned Aerial Vehicle (UAV) flight data; the approach is based on comparing hand-crafted descriptors of UAV behavior (including state, flight plan, and sensor readings) to predefined behavioral profiles.
Park et al. tackle anomaly detection in robot-assisted feeding, using Hidden Markov Models with hand-crafted features~\cite{park2016multimodal}, or a combination of Variational Autoencoders and LSTM networks~\cite{park2018multimodal}; the latter approach encodes a sequence of multi-modal sensory signals into a latent space, then estimates the expected distribution of the received inputs: an anomaly is detected when the negative log-likelihood of the current input, given an expected distribution, exceeds a certain threshold.
\paragraph*{On high-dimensional data}
Semantic anomaly detection on high-dimensional sensing data is used in many different robotics application domains.  Early attempts concerned with autonomous patrolling~\cite{chakravarty2007anomaly} apply image matching algorithms between observed data and large databases of normal images, to identify unexpected differences.
Christiansen et al.~\cite{christiansen2016deepanomaly}, detect obstacles and anomalies on an autonomous agricultural robot, using DeepAnomaly, a custom CNN derived from AlexNet~\cite{krizhevsky2012imagenet}; anomalies are found and highlighted using background subtraction over high-level features of the CNN. 
Wellhausen et el.~\cite{wellhausen2020safe} achieve safe navigation for a legged ANYmal~\cite{hutter2016anymal} robot in unknown environments, avoiding footholds located on terrain whose appearance is anomalous with respect to the robot's previous experience; the paper compares a standard autoencoder with two additional models sharing the same encoder: Deep SVDD~\cite{ruff2018deep} and Real-NVP~\cite{dinh2016density}; Real-NVP performed best, followed closely by the autoencoder.

\subsection{Anomaly Detection Datasets}
\label{sec:rel_data}
Most Anomaly Detection literature~\cite{hendrycks2018deep, ruff2018deep, tuluptceva2019perceptual} relies\mdiff{,} for experiments\mdiff{,} on datasets built for image classification, including MNIST~\cite{lecun1998gradient}, ImageNet~\cite{deng2009imagenet,le2015tiny}, CIFAR~\cite{krizhevsky2009learning}\mdiff{,} and SVHN~\cite{netzer2011reading}.  In particular, in this context, one can define an anomaly as an instance sampled from a different dataset than the one used for training; or an instance of a given class that is not seen in the training set.

In robotics applications like the scenario we focus on, this is not a good model for realistic anomalies: in fact, anomalous frames might differ from normal ones in subtle ways. To properly develop and evaluate anomaly detection methods for these applications, researchers need datasets that represent \emph{realistic} anomalies, as well as large amounts of normal images with their expected variability.  Some recently-released datasets tackle this issue and focus explicitly on Anomaly Detection~\cite{blum2021fishyscapes,wellhausen2020safe,bergmann2021mvtec}.

MVTec~\cite{bergmann2021mvtec} is specific to defects of industrial goods; it is composed of 5354 RGB images of 15 different objects captured in controlled scenarios in an industrial production environment. 
Fishyscapes~\cite{blum2021fishyscapes}, built upon Cityscapes~\cite{cordts2016cityscapes}, focuses on segmenting anomalous objects in images acquired by self-driving vehicles. It includes both anomalous images acquired in real-world settings 
as well as synthetically-generated data; dense segmentation masks of anomalous objects \mdiff{are provided}.
Wellhausen et al.~\cite{wellhausen2020safe} propose ANNA, a dataset for detecting visual anomalies in ground patches, to predict terrain traversability for legged robots; the dataset is built from observations acquired by robots traveling on different terrains.

\begin{figure*}[!t]
    \centering
    \begin{subfigure}[t]{0.3\textwidth}
    \centering
    \includegraphics[height=4cm]{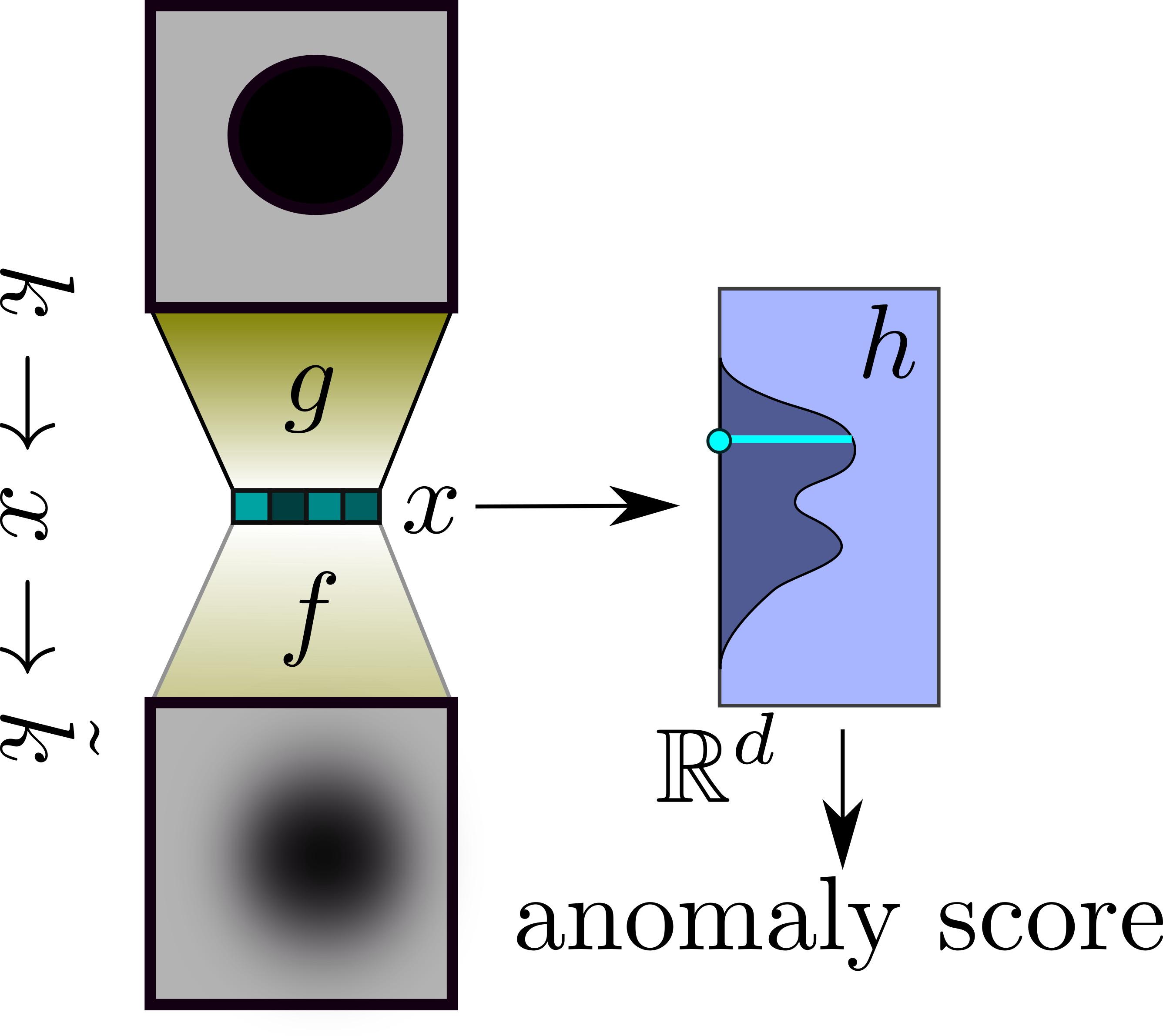}
    \caption{\emph{Anomaly detector}: the anomaly score is computed from the estimated likelihood of the encoding of image $k$. The auto-encoder is trained to reproduce $\tilde k\approx k$.}
    \label{fig:ad}
    \end{subfigure}\hfill
     \begin{subfigure}[t]{0.305\textwidth}
    \includegraphics[height=4cm]{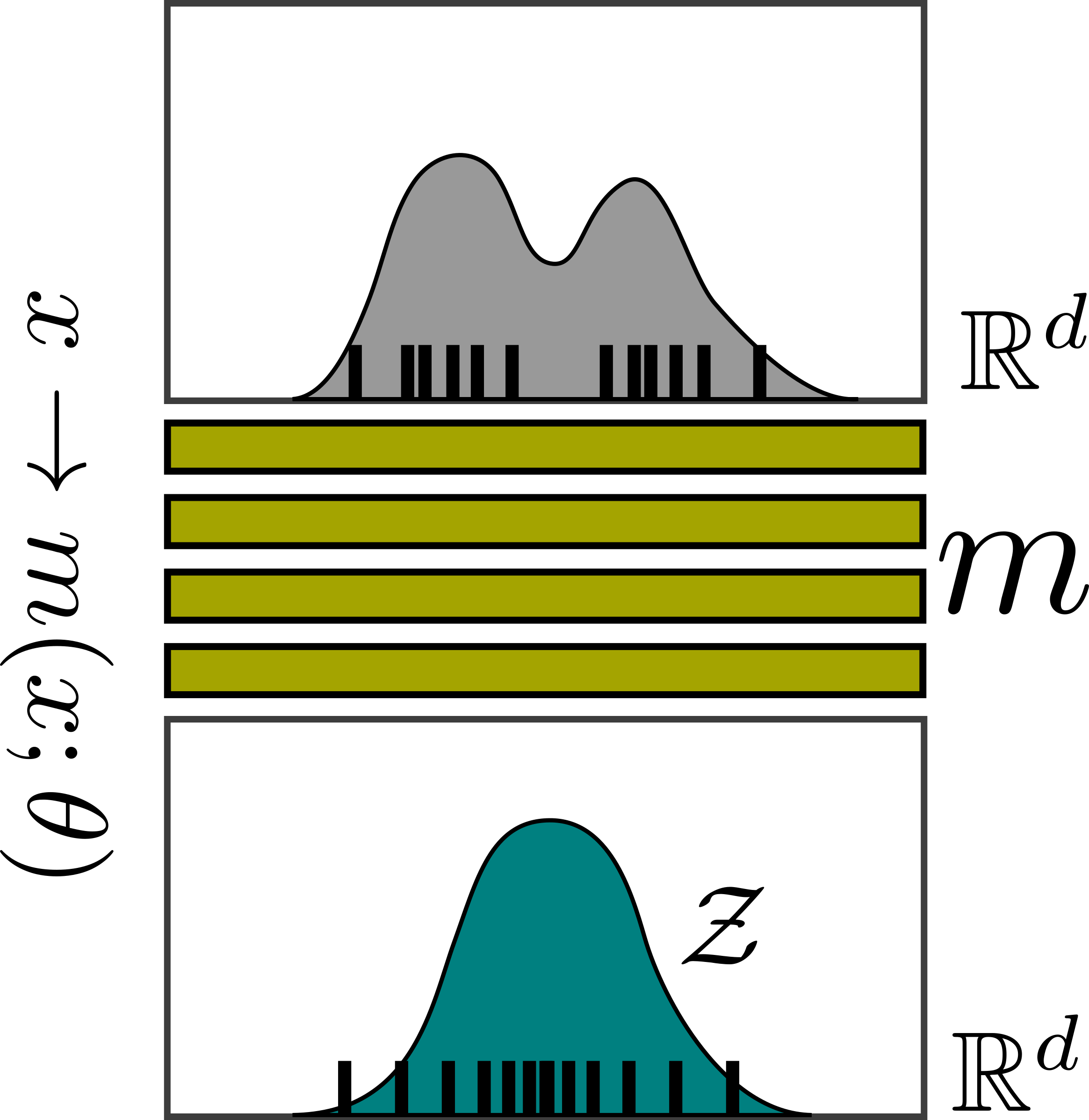}
    \centering
    \caption{\emph{Real-NVP} learns the best model $m$ that maximizes samples' likelihood by mapping them to a target distribution $\mathcal{Z}$ through a series of invertible coupling layers.}
    \label{fig:rnvp}
    \end{subfigure}\hfill
    \begin{subfigure}[t]{0.345\textwidth}
    \includegraphics[height=4cm]{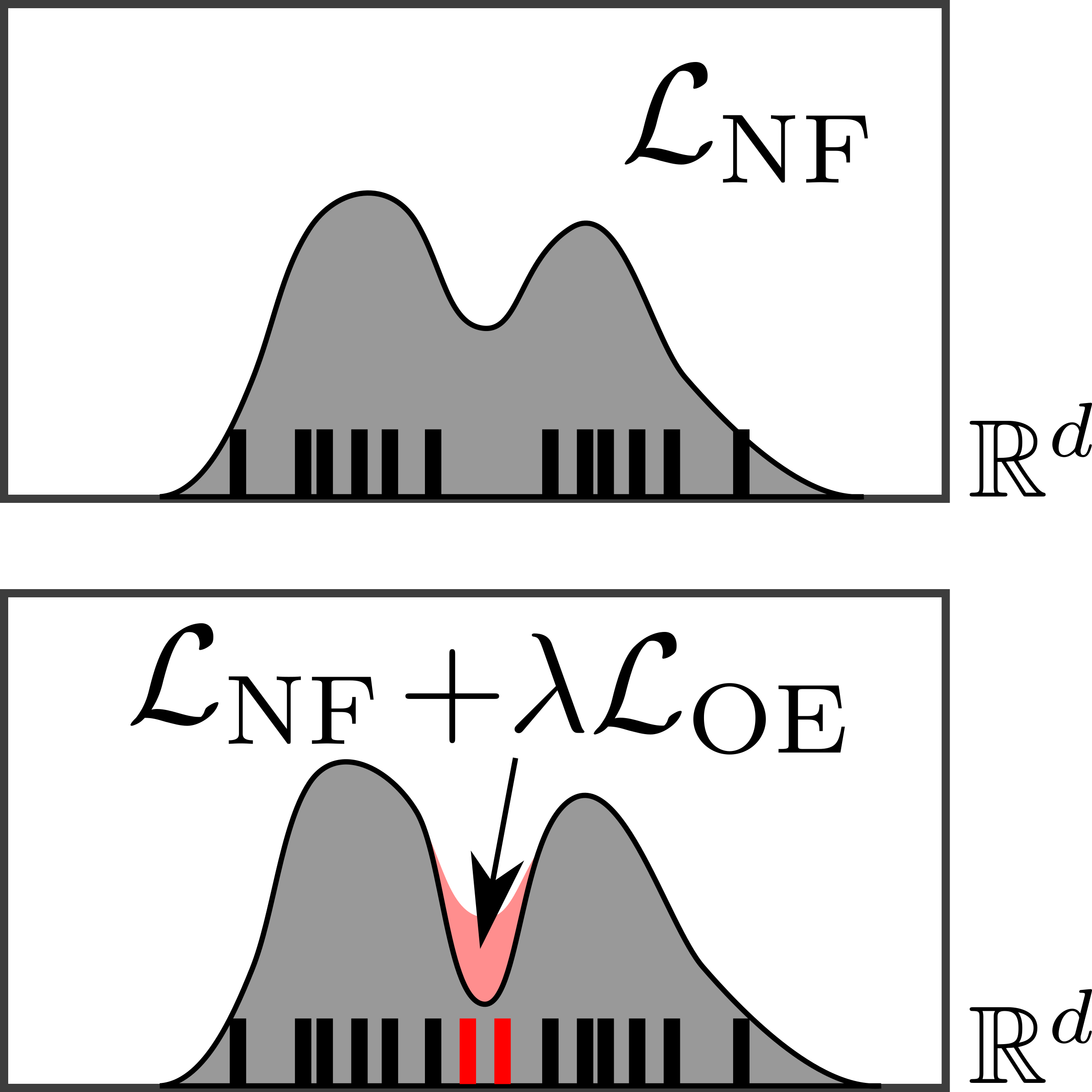}
    \centering
    \caption{\emph{Outlier exposure}: (above) the original Real-NVP loss does not consider anomalous samples; (below) adding an outlier exposure loss reduces the likelihood assigned to known anomalies.}
    \label{fig:rnvp+oe}
    \end{subfigure}
    \label{fig:method}
    \caption{The method to compute anomaly scores presented in Section~\ref{sec:method}. \mdiff{Gray} plots in (b) and (c) represent probability densities over the encoding space, where dataset samples are drawn as small rectangles (normal in black, anomalous in red)}
\end{figure*}

\section{Method}\label{sec:method}


\subsection{Problem statement and definitions}

Our goal is to decide, at run-time, whether the current camera frame $k$ is anomalous or not.
We do this by training machine-learning models on a dataset $\mathcal{K}$ of previously acquired images. The dataset is composed of two disjoint subsets: $\mathcal{K}_n$, with images labeled as normal, and $\mathcal{K}_a$, generally much smaller, with images labeled as anomalies.

The anomaly detector takes as input an image and outputs a real-valued anomaly score. Following a common approach~\cite{wellhausen2020safe},
the anomaly detector consists of two parts: dimensionality reduction and density estimation (see \mdiff{Fig.}~\ref{fig:ad}).

\paragraph{Dimensionality reduction}
We train an autoencoder $f \circ g$ to reconstruct images in $\mathcal{K}_n$.  Once trained, we use the encoder to compute lower-dimensional representations $x = g(k) \in \mathbb{R}^d$. In particular, we obtain two datasets  $\mathcal{X}_n = g(\mathcal{K}_n)$ and  $\mathcal{X}_a = g(\mathcal{K}_a)$.
 Adding samples to $\mathcal{K}_a$ does not require re-training the encoder; in fact, for this paper, the encoder has been trained once and then shared by the different anomaly detectors.

\paragraph{Density estimation}
A second machine-learning model $h(\cdot; \theta)$
estimates the probability density of the distribution $\mathcal{D}_n$ from which $\mathcal{X}_n$ has been sampled.
Parameters $\theta$ are trained on both $\mathcal{X}_n$ and $\mathcal{X}_a$.

Once both models have been trained, we define the anomaly score of an image $k$ as the negative log-likelihood of its encoding, assuming it belongs to $\mathcal{D}_n$, i.e.:
\begin{equation}
\label{eq:ad}
-\log h(g(k); \theta) \in \mathbb{R}\,.
\end{equation}

We devote the rest of the section to describe the density estimator $h$: a variant of Real-NVP that additionally takes known anomalies into account.

\subsection{Real-NVP}
\label{sec:rnvp}
\mdiff{Normalizing flow models} estimate the density of a distribution $\mathcal{D}_n$ from a set of samples. They are based on learning the parameters $\theta$ of a bijection $m (\cdot; \theta): \mathbb{R}^d \to \mathbb{R}^d$ that maps $\mathcal{D}_n$ to a known target distribution $\mathcal{Z}$. Following a change of variables, the probability density of the original samples is recovered from the target probability density as:
\begin{equation}
    \label{eq:change_of_variables}
    p(x| \theta) = p_{\mathcal{Z}}\left(m(x; \theta)\right)\left|\det \left(\frac{\partial m(x; \theta)}{\partial x} \right) \right|\,.
\end{equation}

For the rest of the paper, we fix  
the target to be a multi-variate normal distribution of zero mean and unit variance $\mathcal{Z} = \mathcal{N}(0, \mathbb{1})$.\footnote{This approach is known as Gaussianization~\cite{chen2000gaussianization} when $\mathcal{Z}$ is a normal distribution.}

Normalizing flow models are trained to maximize the likelihood of the samples in $\mathcal{X}_n$, or equivalently to minimize their average negative-log-likelihood loss function:
\begin{small}
\begin{eqnarray}
    \label{eq:lRNVP}
    \mathcal{L}_{\mathrm{NF}}(x) &=& -\log(p(x| \theta)) \\
    \nonumber
    &=&-\log(p_{\mathcal{Z}}(m(x; \theta))) -\log \left(\left|\det \left(\frac{\partial m(x; \theta)}{\partial x} \right) \right| \right)\,.
\end{eqnarray}
\end{small}
While the first term is easily computed, and equals to 
$\frac{d}{2}\log(2 \pi) + \frac{1}{2} |m(x)|^2$ for the chosen $\mathcal{Z}$, clever solutions are required to keep the second term tractable.
We make use of one of them: Real-NVP, a deep-learning normalizing flow model~\cite{dinh2016density}, where $m$ consists of a sequence of \emph{affine coupling layers} designed to 
simplify the computation of the determinant of the derivative (see \mdiff{Fig.}~\ref{fig:rnvp}). Each coupling layer contains parametric translation and scaling on a subspace of $\mathbb{R}^d$, modeled as a dense neural \mdiff{network}.

Once the model is trained, we can infer the negative-log-likelihood of a sample, which is then used as an anomaly score:
\begin{equation}
    \label{eq:anomaly_score_RNVP}
    -\log h(x) = \mathcal{L}_{\mathrm{NF}}(x)\,.
\end{equation}

\subsection{Real-NVP with outlier exposure}
\label{sec:rnvp+oe}
Density estimation through Real-NVP makes no use of known anomalous samples. To solve this limitation, we combine it with outlier exposure.

The goal of outlier exposure is to improve the performance of an anomaly detection model $m$ by exposing it to samples $\mathcal{X}_a$ drawn from an outlier distribution $\mathcal{D}_a$, using an additional loss $\mathcal{L}_{\textrm{OE}}$ applied to pairs of samples $(x_n, x_a) \in \mathcal{X}_n \times \mathcal{X}_a$. In its generic form, the model should now minimize a weighted sum of the original loss (in our case $\mathcal{L}_{\textrm{NF}}$) and of $\mathcal{L}_{\textrm{OE}}$:
\begin{equation}
\label{eq:oeog}
\langle \mathcal{L_{\textrm{NF}}}(x_n) \rangle_{ \mathcal{X}_n}
+\lambda \langle \mathcal{L}_{\textrm{OE}}(x_n,x_a)\rangle_{ \mathcal{X}_n \times \mathcal{X}_a}\,,
\end{equation}
where $\langle \cdot \rangle_\mathcal{X} $ denotes the average over a set $\mathcal{X}$.

Following~\cite{hendrycks2018deep}, we define $\mathcal{L}_{\mathrm{OE}}$ as a margin ranking loss:
\begin{equation}
    \label{eq:margloss}
    \mathcal{L}_{\mathrm{OE}}(x_n, x_a)=\max\left(0,\gamma + \mathcal{L}_{\mathrm{NF}}(x_n)-\mathcal{L}_{\mathrm{NF}}(x_{a})\right)\,,
\end{equation}
where $\gamma$ represents the margin hyperparameter; $\mathcal{L}_{\mathrm{OE}}$ is added to  $\mathcal{L}_{\mathrm{NF}}$ loss
as in Eq.~\eqref{eq:oeog}.
The resulting loss, besides maximizing the likelihood of normal samples, further encourages the model to assign a higher likelihood to normal samples than to known anomaly samples 
(see \mdiff{Fig.}~\ref{fig:rnvp+oe}).

\begin{figure}[t]
    \centering
    \includegraphics[trim={0 10mm 0 5mm},clip, width=1.0\columnwidth]{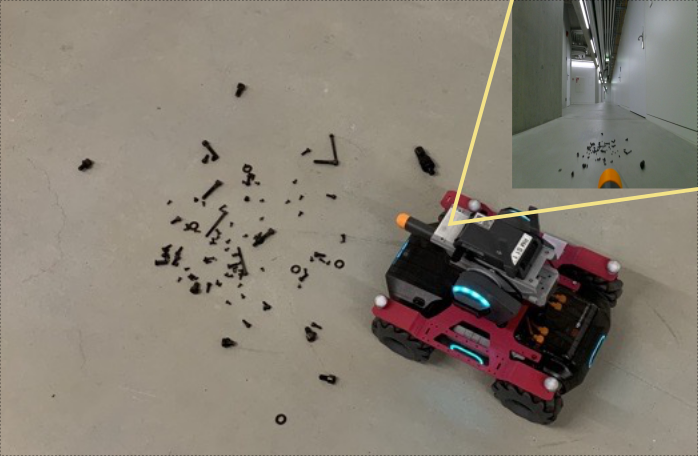}
    \caption{Our dataset is collected using a Robomaster S1 ground robot, patrolling an indoor environment.}
    \label{fig:RM}
\end{figure}

\begin{figure*}[tb]
    \centering
    \includegraphics[width=\textwidth]{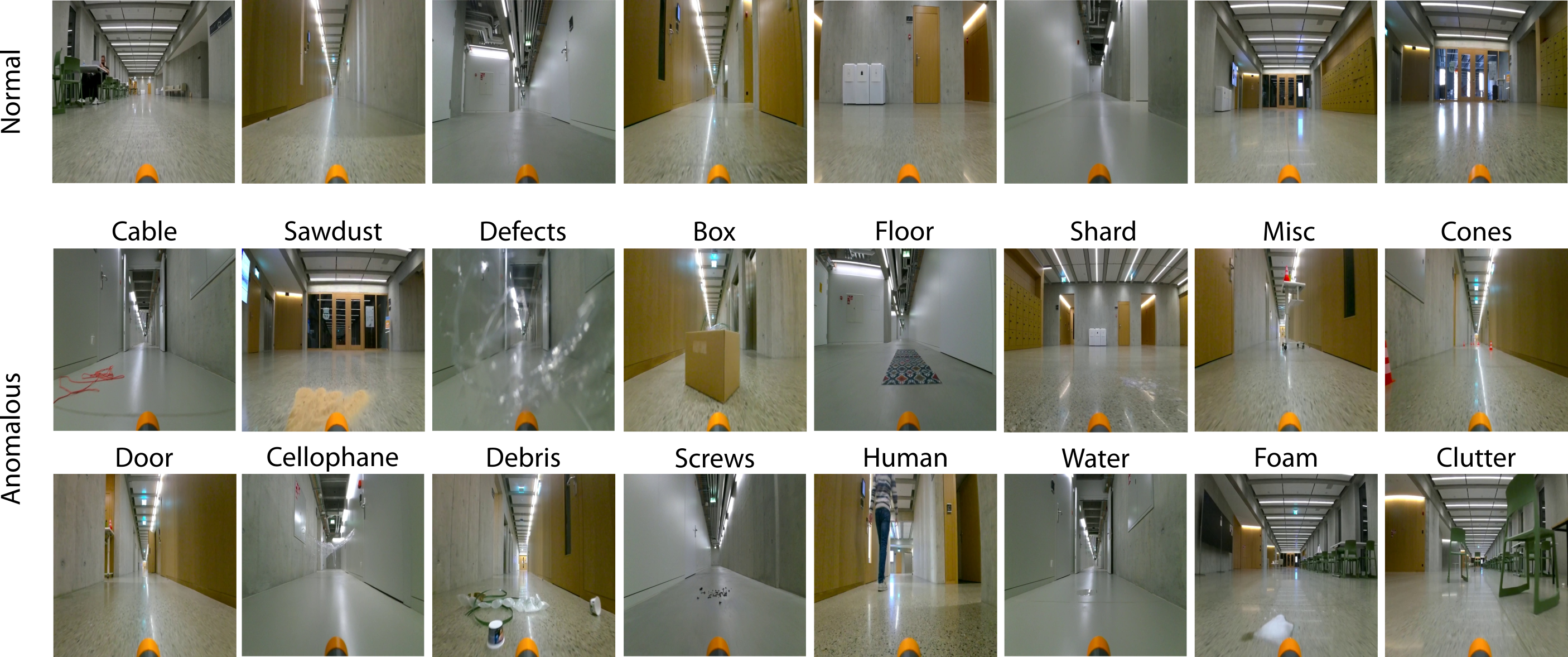}
    \caption{First row: a random sample of normal frames.  Second and third rows: examples of all 16 anomaly types.}
    \label{fig:dataset}
\end{figure*}

\section{Experimental setup}
\label{sec:experimental_setup}
\subsection{Dataset}
\label{sec:dataset}

\mdiff{
We test the approach described in Section~\ref{sec:method} on an indoor scenario that is relevant for diverse tasks, such as safe navigation (where anomalies are potential hazards to avoid), or patrolling (where anomalies are interesting events to report); we study anomaly detection independently of the specific task that the robot may be doing. 

The dataset is composed of frames, each of which is labeled as either normal or as representing a certain class of anomaly.  It has been acquired by the front-facing camera of an omni-directional wheeled robot (Robomaster S1, see Fig.~\ref{fig:RM}) teleoperated along corridors of office and university buildings, featuring different characteristics, with both natural and artificial illumination. In normal situations, the robot travels along tidy, empty corridors. 
Most anomaly instances were intentionally caused by us during data acquisition; others happened by chance (such as the unexpected presence of people or misplaced objects).

Anomaly classes cover different categories of anomalies, such as global image artifacts (e.g., \emph{defects} like defocused or misaligned frames) and semantic anomalies (e.g., \emph{doors} labels open doors in a context where doors should always be closed). Classes are visually diverse as possible: some very subtle (e.g., small \emph{screws} on the floor), other are very prominent (e.g., \emph{humans} passing by). Some represent direct hazards to the robot (e.g., \emph{water} puddles to avoid), while others are harmless (e.g.,  hanging \emph{cellophane} pieces). Finally, some anomalies may be predicted and addressed by common algorithms (e.g., \emph{boxes}), while others represent unpredictable, very unlikely situations (e.g., \emph{saw dust} in indoor corridors). 

The dataset is composed of 132838 $512 \times 512$ RGB frames: 105122 labeled as normal and 27716 labeled as one of 16 types of anomalies (see Fig.~\ref{fig:dataset}).\footnote{The complete dataset, as well as a preliminary version presented in previous work~\cite{mantegazza2021sensing}, is available at \url{{https://github.com/idsia-robotics/hazard-detection}}; it contains two additional scenarios of drones flying in indoor environments, which are not used in this paper.
}
The dataset is split into disjoint training, validation, and testing sets, as shown in Table~\ref{tab:dataset-split}; the testing set originates from different sequences than the training set.  
%
For training, the type of the anomaly is ignored: they are all treated as one class, therefore the model outputs an anomaly score without predictions about the type. The type of anomalies is only used to compare the model performance on different anomalies.
}

\begin{table}[h]
\centering
\caption{Dataset split}
\label{tab:dataset-split}
\begin{tabular}{@{}rccc@{}}
\toprule
                          & training set & validation set & testing set \\ \midrule
normal frames    & 56066                & 2471                   & 46585               \\
anomalous frames & 18201                  & -                       & 9515                \\ \bottomrule
\end{tabular}
\end{table}
\subsection{\mdiff{Autoencoder: reconstruction and dimensionality reduction}}
\mdiff{We train an undercomplete autoencoder to detect anomalies via the reconstruction method (AE)}.
The autoencoder implements a simple convolutional architecture often used for similar purposes~\cite{geng2015high,guo2017deep}: it takes as input a $64\times64$ RGB image; compresses it to a 128-dimensional representation; then reconstructs it as a $64\times64$ RGB image as output.
The model is trained on normal samples of the training set by minimizing the mean squared error loss between input and output. \mdiff{At inference time, the mean squared reconstruction error of a given frame is used as an anomaly score.} 

\mdiff{The same autoencoder is also used for dimensionality reduction: the trained encoder $g$ extracts 128-dimensional visual features from input images, which are used by the approaches described in Section~\ref{sec:de} (see Fig.~\ref{fig:ad}).}

\begin{figure}[t]
    \centering
    \includegraphics[width=\columnwidth]{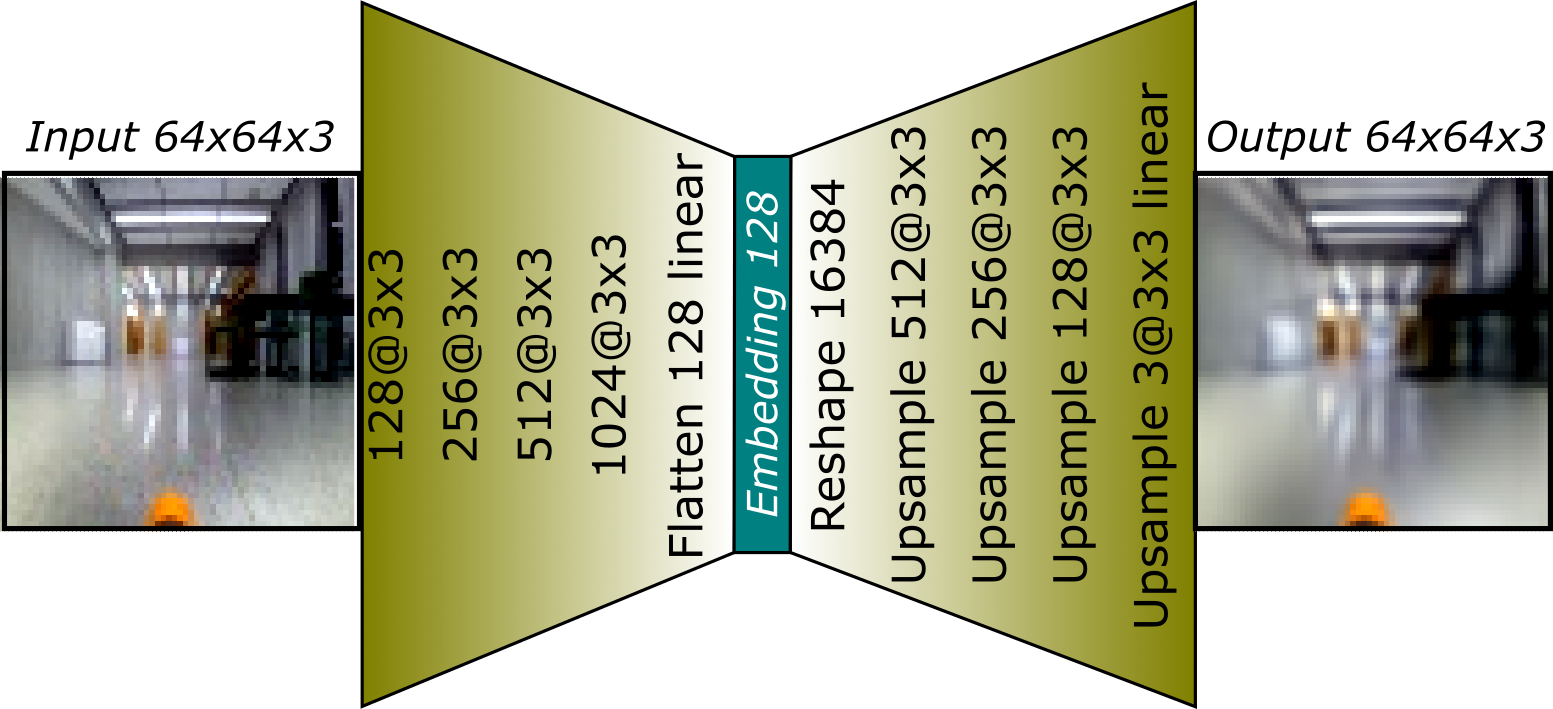}
    \caption{Autoencoder architecture.
    Encoder layers (left) use a stride of two. Inputs of decoder layers (right) are up-scaled before convolutions. When not specified, layers use a LeakyReLU activation function.}
    \label{fig:AE}
\end{figure}

\mdiff{Fig.~\ref{fig:AE} illustrates the model architecture: both the} encoder and the decoder consist of four convolutional layers interleaved by LeakyReLU activation functions; the last activation function of the decoder is linear. The encoder convolution layers have stride 2 thus halving the input size, while before each decoder convolutional layer the input size is doubled. The bottleneck module is composed of two dense layers that take the output of the encoder and compress it to a 128-dimensional vector.

The autoencoder is trained for 100 epochs using the Adam~\cite{kingma2014adam} optimizer, with a learning rate of 0.001 and a reduction factor of 10 in case of a validation loss plateau. During training, samples are randomly augmented with horizontal flipping, rotation ($\pm 10^\circ$), cropping, contrast and brightness variations, and mild additive Perlin noise~\cite{perlin1985image}. The augmentation pipeline is provided as supplementary material.

\subsection{Density estimation (RNVP, RNVP+OE)}
\label{sec:de}

We detect anomalies using Real-NVP density estimators; we separately report results without outlier exposure (RNVP, Section~\ref{sec:rnvp}) and with outlier exposure (RNVP+OE, Section~\ref{sec:rnvp+oe}). 
Both models share the same architecture\mdiff{:
they} take a 128-dimensional image encoding as input (see \mdiff{Fig.}~\ref{fig:rnvp}), and output a 128-dimensional vector. 
%
The models are composed of four coupling layers whose scaling and translation modules have a single hidden layer with 128 neurons, using \emph{odds} input masking.

The RNVP model is trained by minimizing the loss defined in Eq.~\ref{eq:lRNVP}; the RNVP+OE model is trained by minimizing the loss defined by Eq.~\ref{eq:oeog} and Eq.~\ref{eq:margloss}, where we set $\lambda=1$ and $\gamma=100$.
For each experimental run, both models are trained from scratch, with randomly initialized weights, for 500 epochs using Adam~\cite{kingma2014adam} with a starting learning rate of 0.001. We reduce the learning rate by a factor of 10 in case the validation loss plateaus for more than 10 epochs. For all models, the checkpoint that minimizes the validation loss (which only accounts for normal samples) is selected.

For inference, both models use the same anomaly score defined in Eq.~\eqref{eq:anomaly_score_RNVP}.



\subsection{Binary Classifier Baseline (BCLASS)}

We 
compare RNVP and RNVP+OE with a neural network binary classifier \mdiff{(}BCLASS\mdiff{)} that 
%
operates on the same 128-dimensional image embedding as input. The model $c$ is composed of three fully connected layers (with 256, 64, and 1 neurons respectively) interleaved by a ReLU activation function; a Sigmoid activation function is used at the output.  

The model is trained on $\mathcal{X}_n\cup\mathcal{X}_a$ using the Binary Cross Entropy loss to estimate whether the input is an anomalous frame.
The rest of the training details are shared with RNVP and RNVP+OE.
For inference, a given frame $k$ is assigned an anomaly score $c(g(k))$ corresponding to the output 
of the trained classifier.

\subsection{Wide Residual Networks with Outlier Exposure (WRN+OE)}
In addition to BCLASS, we also compare to Wide Residual Networks (WRN\mdiff{+OE})~\cite{zagoruyko2016wide}, leveraging an architecture of size 16-2 and the implementation used by Hendricks et al. \cite{hendrycks2018deep} for the SVHN dataset~\cite{netzer2011reading}, with the hyperparameters and augmentation used for training the WRN on Tiny ImageNet~\cite{le2015tiny}. To make the approaches comparable, in the training and validation sets we included a new 3-class label representing the environment where the samples were recorded; this label is used only during training and is ignored for testing. For inference, we use the Maximum Softmax Probability (MSP)~\cite{hendrycks2016baseline} over the WRN's output to detect anomalies~\cite{hendrycks2018deep}.  The approach is thus trained to classify which of the three environments each sample was acquired in; the uncertainty in the prediction is used as an anomaly score of the sample.

\subsection{Metrics}
\label{sec:metric}

We evaluate each model on the same testing set, which is composed of both normal and anomalous samples.  The predictive performance of anomaly detection models is measured~\cite{chandola2009anomaly} by its Area Under the ROC Curve (AUC), a robust metric that is independent of the choice of the threshold.  The AUC value can be interpreted as the probability that a random anomalous sample is assigned a higher \mdiff{anomaly} score than a random normal sample. An AUC value of 1.0 corresponds to a perfect anomaly detector that assigns a higher score to anomalies than to any normal sample.

\subsection{Computational costs}
Experiments are run on an NVIDIA 2080 Ti using Python~3.8 and PyTorch~1.7.1. On this platform, training time amounts to 100 minutes for training AE, 20 minutes for training the RNVP model, and BCLASS, 60 minutes for WRN+OE (over 20 epochs but converged after 4), and on average 30 minutes for RNVP+OE. 
Inference time amounts to \SI{1}{ms} for BCLASS, \SI{2}{ms} for WRN+OE, and \SI{5}{ms} per frame for both RNVP and RNVP+OE.

\section{Results}
\label{sec:experimental_results}

In this section, we report the results of an experimental investigation on the performance of RNVP+OE on the dataset described in Section~\ref{sec:dataset}. 
\mdiff{As an initial step, we encoded all images in the dataset} using the autoencoder's encoder,
which yields a dataset of embeddings; this dataset is used in all RNVP, BCLASS, and RNVP+OE experiments.
\subsection{Hyperparameter exploration}

We explored various hyperparameters of autoencoder, RNVP, BCLASS, WRN, and RNVP+OE. Most of the hyperparameters were set using a preliminary search or were inspired by previous research~\cite{wellhausen2020safe,dinh2016density,hendrycks2018deep, mantegazza2021sensing}: the autoencoder's bottleneck size, learning rate, maximum number of epochs, input size and architecture details;  the RNVP input size, learning rate, coupling layer size and number, and input masking; WRN's architecture and hyperparameters.
For all experiments except AE and WRN+OE, we let the model run for 500 epochs, then choose the best performing model over the validation set. For WRN+OE we observed that after 4 epochs the model converged, thus we let the model train for at most 20 epochs with early stopping.

\begin{figure}[tb]
    \centering
    \includegraphics[trim=25 0 0 0, clip,width=\columnwidth]{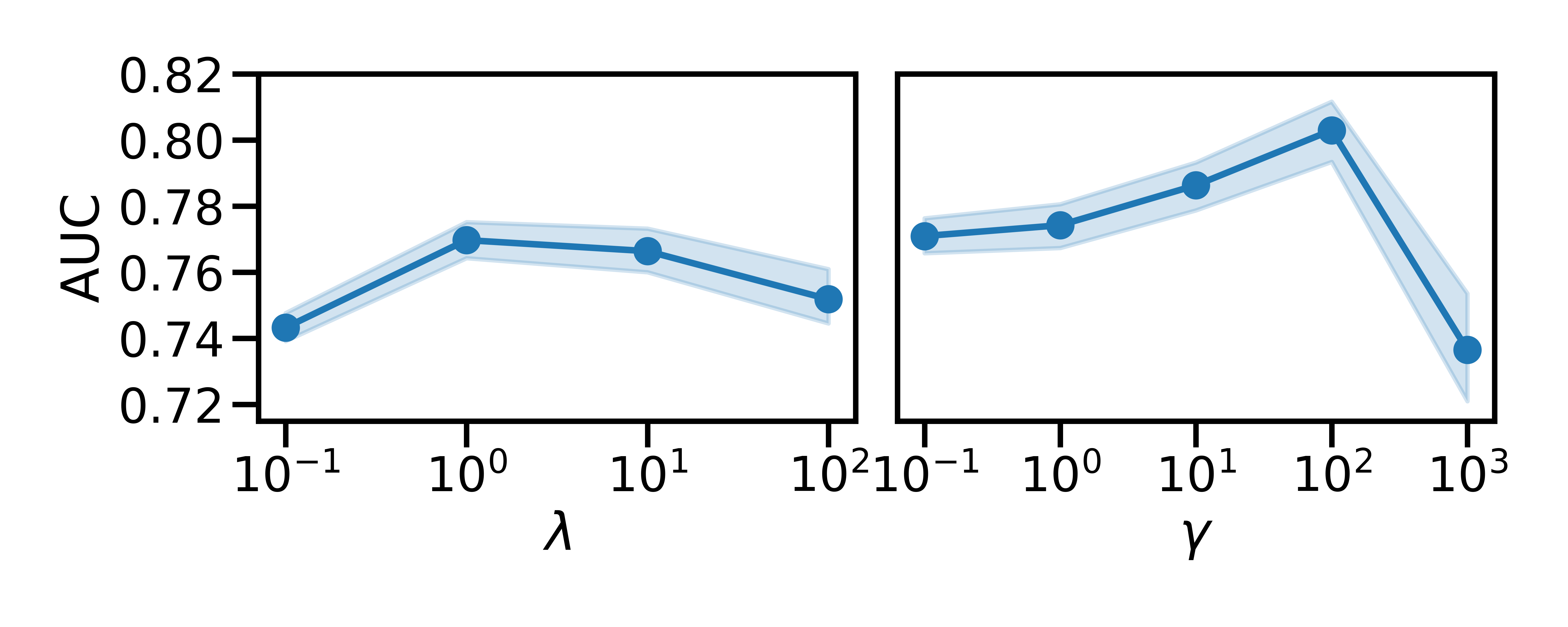}
    \caption{The impact of parameters $\lambda$ and $\gamma$}
    \label{fig:an4-5}
\end{figure}

\mdiff{Fig.}~\ref{fig:an4-5} illustrates how we selected parameters $\lambda$ and $\gamma$ in Eq.~\eqref{eq:margloss} and Eq.~\eqref{eq:oeog}, which are specific to our proposed contribution.
\mdiff{Fig.}~\ref{fig:an4-5}:left shows the impact of $\lambda$, for fixed $\gamma=0$, on 10 experiments for each value: we select $\lambda=1$ as the best parameter.
Similarly, \mdiff{Fig.}~\ref{fig:an4-5}:right shows the impact of $\gamma$ for $\lambda=1$ on 10 experiments for each value:  we select $\gamma=100$ as the best parameter.

\subsection{Comparison with baselines}
We compare the performance of RNVP+OE with the baselines AE, WRN+OE, RNVP and BCLASS.
RNVP+OE, WRN+OE and BCLASS are trained on the full training set, RNVP only on the normal samples of the training set and AE on the normal frames; all are tested on the same testing set (AE is tested on the corresponding frames). For RNVP, RNVP+OE, and BCLASS, we run the experiment 10 times to reduce the noise originating from the stochastic training of the models. AE and WRN+OE are trained only once due to the greater computation requirements. For the AE we report two AUCs based on two different anomaly scores, Mean Absolute Error (MAE) and Mean Squared Error (MSE) over the reconstruction error.

\begin{table}[!h]
\centering
\caption{AUC values for our model and baselines (* indicates average over 10 runs)}
\label{tab:an1_auroc}
\begin{tabular}{@{}cccccc@{}}
\toprule
\multicolumn{3}{c}{no OE}  & \multicolumn{3}{c}{with OE} \\ \cmidrule(r){1-3}\cmidrule(l){4-6}
AE-MAE & AE-MSE & RNVP & BCLASS  & WRN+OE  & RNVP+OE\\
0.64 & 0.67 & 0.73*   & 0.73*   & 0.55 & \textbf{0.80*}
 \\ \bottomrule
\end{tabular}   
\end{table}

\begin{figure}[tb]
    \centering
    \begin{subfigure}[c]{0.95\columnwidth}
    \centering
    \includegraphics[trim=25 220 25 0, clip, width=\textwidth]{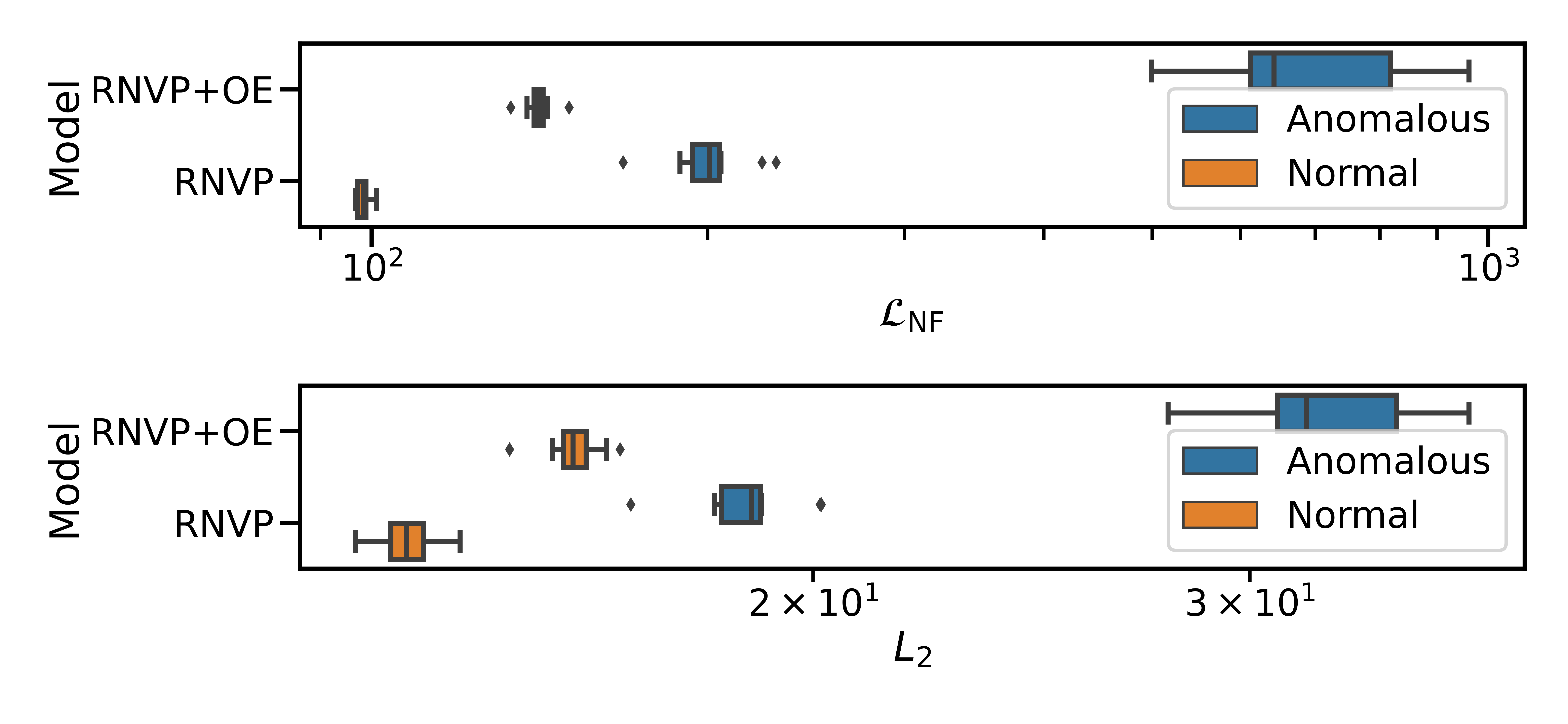}
    \caption{The loss function $\mathcal{L}_{\mathrm{NF}}$, i.e., the negative-log-likelihood, evaluated over the testing set. Larger values mean less likely.}
    \label{fig:loss_an1}
    \end{subfigure}
    \par\bigskip
    \begin{subfigure}[c]{0.95\columnwidth}
    \includegraphics[trim=25 0 25 220, clip, width=\columnwidth]{img/analysis/an1_boxplotv2.png}
    \caption{The euclidean norm of transformed samples $m(x)$ evaluated over the testing set.}
    \label{fig:l2_an1}
    \end{subfigure}
    \caption{The effect on loss function and $L_2$ norm of exposing the model to anomalies during training.}
    \label{fig:dist_an1}
\end{figure}

\paragraph*{Results}
Table~\ref{tab:an1_auroc} reports the AUC of all baselines. WRN+OE (AUC=0.55) yields a low performance on the task, which is probably caused by the fact that many considered anomaly types do not hinder the prediction of the environment in which the sample is acquired.  Despite not using exposed outliers for training, AE yields a better performance both when using the reconstruction MAE (AUC=0.64) and MSE (AUC=0.67) as anomaly scores.
 BCLASS (AUC=0.73) and RNVP (AUC=0.73) show promising performance on the dataset. Finally, RNVP+OE has a significantly better performance than all alternatives (AUC=0.80).
\mdiff{\subsection{Effect on the RNVP target space}
We investigate the effect of exposing RNVP to anomalies.
\paragraph*{Results}
Fig.~\ref{fig:dist_an1} shows how
the additional loss defined in Eq.~\eqref{eq:margloss} increases the negative-log-likelihood of anomalous testing samples, leading to a wider separation from normal samples (Fig.~\ref{fig:loss_an1}).
Using outlier exposure on Real-NVP, anomalous samples are mapped further away from the origin (i.e., the mean of the target multivariate normal distribution) by RNVP+OE than by RNVP (Fig.~\ref{fig:l2_an1}).
}

\subsection{Impact of the number of exposed anomaly frames}
\mdiff{
Our contribution targets situations where limited anomaly samples are known. Therefore,}
we measure how much the number of anomaly samples available at training impacts the performance of RNVP+OE and BCLASS on the testing set\mdiff{, starting from just a few samples}.
\mdiff{We test exposed outlier set sizes ranging from 2 to 16384 samples}; for each size, we train both models 10 times: for each run, we randomly sample a subset of the corresponding size from the exposed outliers in the training set.
Note that RNVP+OE trained on 0 anomalies is equivalent to RNVP.

\begin{figure}[!htb]
    \centering
    \includegraphics[width=\columnwidth]{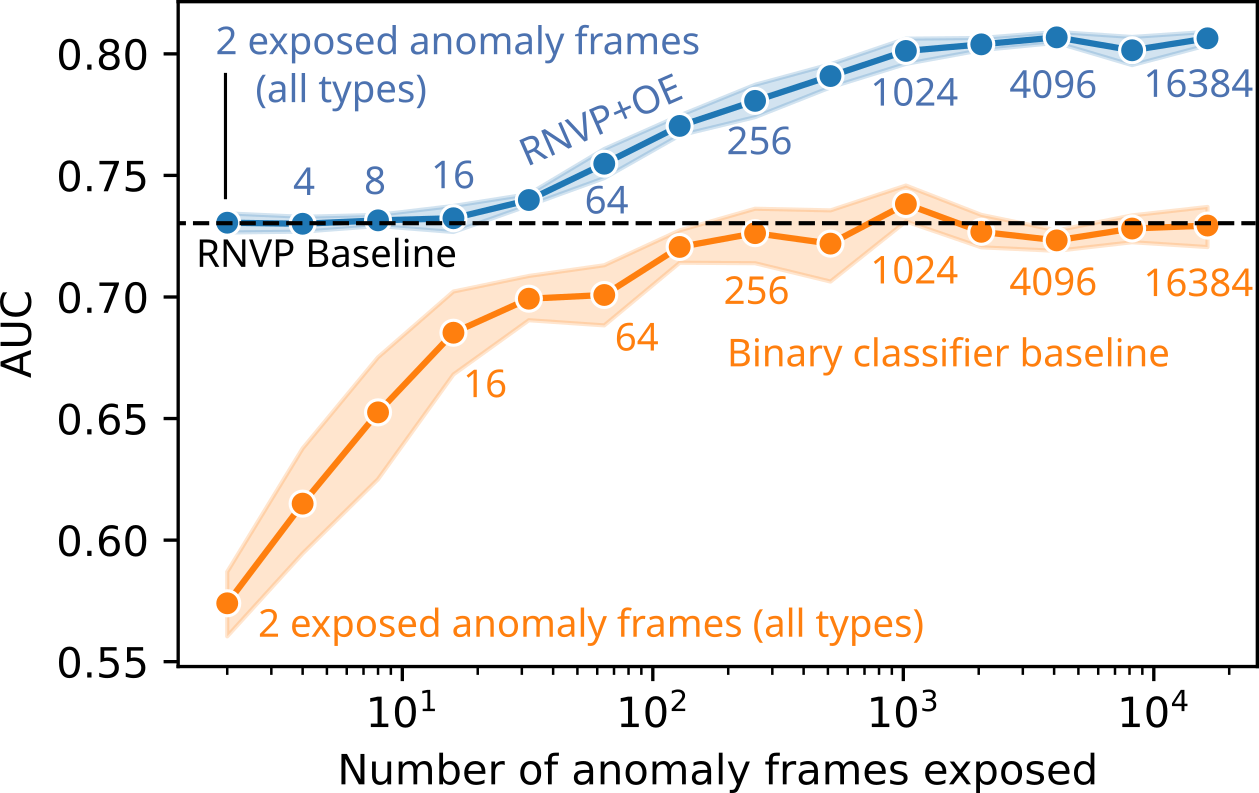}
    \caption{
    Impact of the number of exposed anomaly frames (x-axis, log scale) on the AUC value (y-axis) for the RNVP+OE model (blue) and the binary classifier baseline (orange); line and shaded area correspond to the mean and 95\% c.i. over 10 runs of each model.}
    \label{fig:hist_an2}
\end{figure}

\paragraph*{Results}
\mdiff{Fig.}~\ref{fig:hist_an2} illustrates that, as expected, AUC increases when training with more exposed anomaly frames. While both models exhibit an increase in performance by seeing more anomalies during training, RNVP+OE always outperforms BCLASS. For RNVP+OE, we observe that the performance tends to saturate after 1024 outlier frames are exposed.
Interestingly, even when RNVP+OE is exposed to just a few anomalies (AUC=0.75 for $N=64$), there is a noticeable increase in performance,\mdiff{while BCLASS performs significantly worse than RNVP for $N\le64$.}

\subsection{Impact of the number of exposed anomaly types}
As described in Section~\ref{sec:dataset}, the dataset set is composed of different types of anomalies.
We want to explore the impact of the heterogeneity of the training set by varying the number of types of anomalies available for training an RNVP+OE model.
Using the complete training and testing sets, for $N=1, 2, 3, 4, 6, 12$ types of anomalies, we run the following experiment 30 times: (1) we randomly pick a subset of $N$ anomaly types;
(2) we train RNVP+OE exposing it only to the selected anomaly types; (3) we compute the AUC over the whole testing set and also over the testing set limited to or excluding the anomaly types used for training.

\begin{figure}[!tb]
    \centering
    \includegraphics[width=\columnwidth]{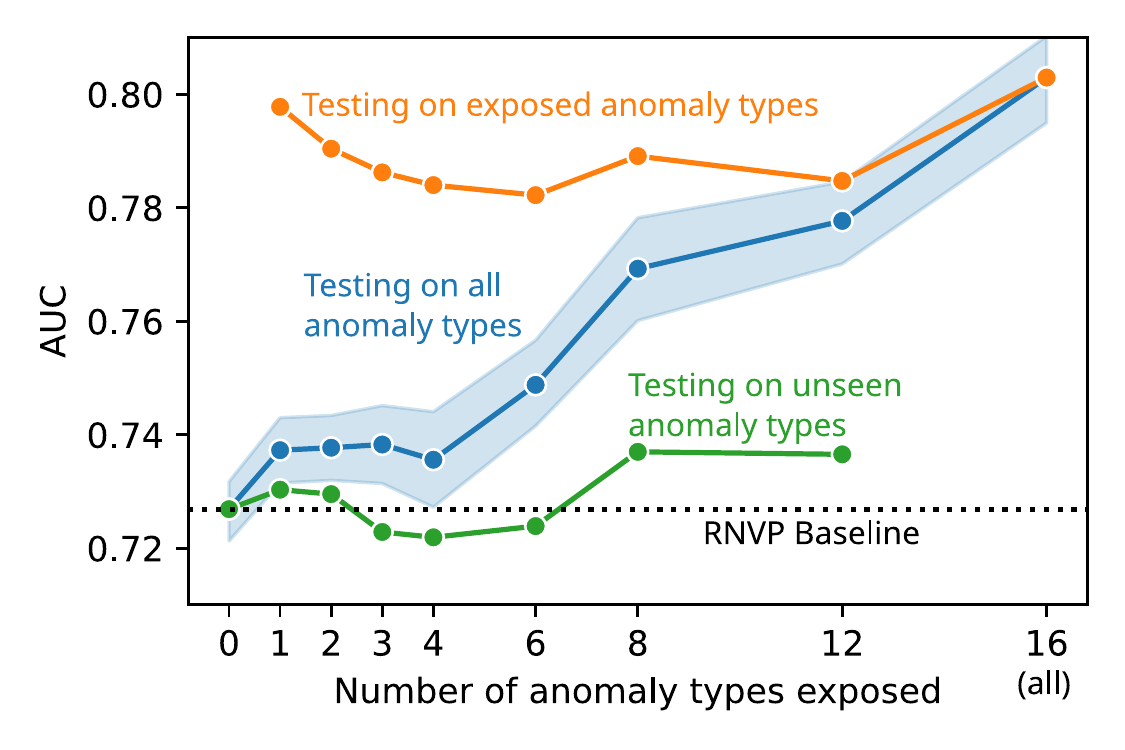}
    \caption{Performance (y-axis) for RNVP+OE models (colored lines), when trained with an increasing number of anomaly types exposed (x-axis).  Performance is reported for: all anomaly types (blue); only the exposed anomaly types (orange); only the not-exposed anomaly types (green). Points correspond to the mean over 10 runs of each model; shaded blue area represents the 95\% c.i..}
    \label{fig:an_temporanea}
\end{figure}

\paragraph*{Results}
\mdiff{Fig.}~\ref{fig:an_temporanea} illustrates how the AUC varies over the different subsets of the testing set (with all anomaly types, only with the anomaly types exposed in training, only with anomaly types not exposed in training).
We observe that exposing additional anomaly types does not hinder the detection of non-exposed anomaly types (green line): we notice a small improvement when more than 6 types are exposed. The most significant improvement is on already seen types of anomalies (orange line), where the performance does not change significantly when the exposed anomalies are more heterogeneous (i.e., more types are exposed). As expected, similar to the previous experiment, the more anomalies are exposed, the better the model performs on the whole testing set (blue line).


\section{Conclusions}
We considered the problem of building visual anomaly detection systems for mobile robots, in situations where a limited amount of examples of some anomalies is available. Our approach combines a Real-NVP model with an additional margin loss term, optimized jointly with the Real-NVP loss.  Quantitative experiments show that exposure to even a small number of anomaly frames significantly improves anomaly detection performance and that the proposed approach significantly outperforms alternative approaches.  We further release a dataset, comprising more than 132k frames collected by a real ground robot in different indoor environments, and representing several occurrences of each of 16 different anomaly types, spanning from sensor defects to high-level semantic anomalies.  \mdiff{Our proposed approach and dataset pave the way for further research aimed at safe autonomous operation of mobile robots in unstructured, unpredictable environments: we foresee applications to concrete robotics tasks (such as avoiding anomalous terrain for legged robots), and extensions of outlier exposure to other anomaly detection approach based on Variational Autoencoders and Generative Adversarial Networks.}

\bibliographystyle{IEEEtran}
\bibliography{biblio}
\end{document}